\documentclass[journal,twoside,web]{ieeecolor}
\usepackage{tmi}
\usepackage{cite}
\usepackage{amsmath,amssymb,amsfonts,bm}
\usepackage{graphicx}
\usepackage{textcomp}
\usepackage{algorithm,algpseudocode,lipsum}

\usepackage{epstopdf} 
\usepackage{tabularx,threeparttable}
\usepackage{array}
\usepackage{mathtools}
\usepackage{mathrsfs}
\usepackage{booktabs}
\usepackage{multirow}
\usepackage{url}
\usepackage{color}
\usepackage{dsfont}
\usepackage{subcaption}
\usepackage{caption}
\usepackage{subcaption}

\def\ie{\emph{i.e.}}
\def\eg{\emph{e.g.}}
\def\etal{{\em et al.}}

\newcommand\T{\rule{0pt}{2.1ex}}       
\newcommand\B{\rule[-0.8ex]{0pt}{0pt}} 
\newcolumntype{?}[1]{!{\vrule width #1}}

\def\BibTeX{{\rm B\kern-.05em{\sc i\kern-.025em b}\kern-.08em
    T\kern-.1667em\lower.7ex\hbox{E}\kern-.125emX}}
\begin{document}
	\title{\textit{DoFE}: Domain-oriented Feature Embedding\\ for Generalizable Fundus Image Segmentation\\ on Unseen Datasets}
\author{Shujun Wang,
	Lequan Yu,
	Kang Li,
	Xin Yang,
	Chi-Wing Fu,~\IEEEmembership{Member,~IEEE}\\
	Pheng-Ann Heng,~\IEEEmembership{Senior Member,~IEEE}
	\thanks{Manuscript  received  January  02,  2020;  revised  June  30,  2020; accepted August 02, 2020. The work described in this paper was supported in parts by the following grants:
		Key-Area Research and Development Program of Guangdong Province, China (2020B010165004),
		Hong Kong Innovation and Technology Fund (Project No. ITS/426/17FP and ITS/311/18FP),
		and National Natural Science Foundation of China with Project No. U1813204.}
	\thanks{S. Wang, K. Li, C.-W. Fu and P.-A. Heng are with the Department of Computer Science and Engineering, The Chinese University of Hong Kong, Hong Kong, China (e-mail: sjwang@cse.cuhk.edu.hk; kli@cse.cuhk.edu.hk; cwfu@cse.cuhk.edu. hk; pheng@cse.cuhk.edu.hk). }
	\thanks{L. Yu is with the Department of Radiation Oncology, Stanford University, Palo Alto, CA 94306, USA (e-mail: lequany@stanford.edu).}
	\thanks{X. Yang is with the School of Biomedical Engineering, Health Science Center, Shenzhen University, Shenzhen, China (e-mail: yangxinknow@gmail.com).}
	\thanks{C.-W. Fu and P.-A. Heng are also with the Shenzhen Key Laboratory of Virtual Reality and Human Interaction Technology, Shenzhen Institutes of Advanced Technology, Chinese Academy of Sciences, Shenzhen, China.}
	\thanks{L. Yu is the corresponding author of this work.}
	\thanks{Copyright \textcopyright~2020 IEEE. Personal use of this material is permitted. However, permission to use this material for any other purposes must be obtained from the IEEE by sending a request to pubs-permissions@ieee.org.}
}


\maketitle

\begin{abstract}
	Deep convolutional neural networks have significantly boosted the performance of fundus image segmentation when {test} datasets {have} the same distribution {as the} training datasets.
	{However, in clinical practice, medical images often exhibit variations in appearance for various reasons, e.g., different scanner vendors and image quality.}
	These distribution discrepancies {could lead the} deep networks to over-fit on {the} training datasets and lack generalization ability on {the} unseen {test} datasets.
	To alleviate this issue, we present a novel Domain-oriented Feature Embedding (\textit{DoFE}) framework to improve the generalization ability of CNNs on unseen target domains by {exploring the} knowledge from multiple source domains.
	Our \textit{DoFE} framework dynamically enriches the image {features} with additional domain prior knowledge learned from multi-source domains to make the semantic {features} more discriminative.
	Specifically, we introduce a Domain Knowledge Pool to learn and memorize the prior information extracted from multi-source domains.
	Then the original image {features are} augmented with domain-oriented aggregated {features}, which {are} induced from the knowledge pool based on the similarity between the input image and multi-source domain images.
	We further design a novel domain code prediction branch to infer this similarity and employ {an attention-guided} mechanism to {dynamically} combine the aggregated {features} with the semantic {features}. 
	We comprehensively evaluate our \textit{DoFE} framework on two fundus image segmentation tasks, including {the} optic cup and disc segmentation and vessel segmentation. 
	Our \textit{DoFE} framework generates satisfying segmentation results on unseen datasets and surpasses other domain generalization and network regularization methods.
\end{abstract}

\begin{IEEEkeywords}
	Optic disc segmentation, optic cup segmentation, {vessel segmentation}, domain generalization, feature embedding
\end{IEEEkeywords}

\section{Introduction}
\begin{figure}[!t]
	\centering
	\includegraphics[width=1.0\linewidth]{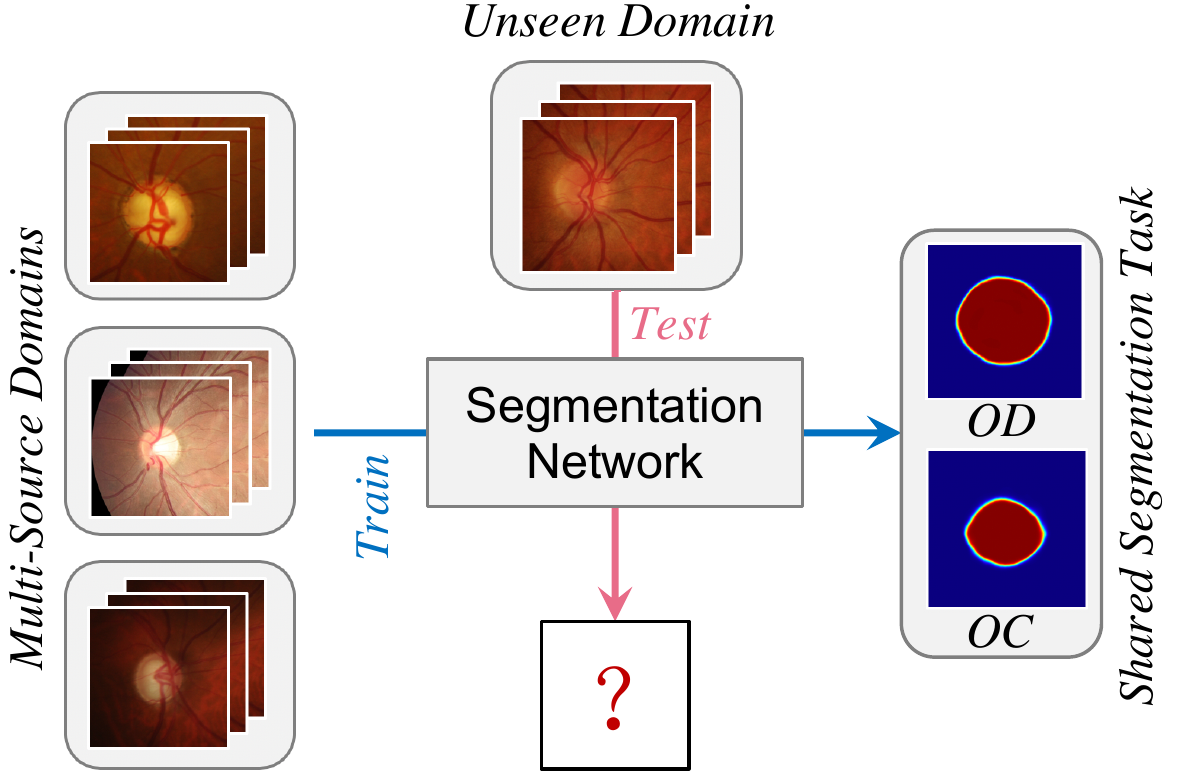}
	\caption{ Illustration of generalizable fundus image segmentation setting. 
		Given the annotated training datasets from multi-source domains, we expect the trained segmentation network to perform well on new images from the unseen target domain without the need of accessing {the} target domain images during the network training.
		OD and OC represent the segmentation masks of optic disc and cup, respectively.
	}
	\label{fig:background}
	\centering
\end{figure}

\IEEEPARstart{R}{etinal} fundus images contain important information for ophthalmic diagnosis.
The structure variation in retinal fundus images is one of the important indicators of certain diseases.
Therefore, automated segmentation of structures, including the optic cup (OC) / disc (OD) and retinal vessel, is critical for clinical diagnosis~\cite{garway1998vertical}.
Deep convolutional neural networks (CNNs) have achieved promising performance in automated OC/OD segmentation and vessel segmentation tasks~\cite{fu2018joint,zhang2019attention}.
However, most of these methods are based on the assumption that {the} training and testing images {have} the same distribution.
{Typically, in real clinical practice, the fundus images are often acquired by different institutions with various types of scanner vendors, patient populations, and disease severity, exhibiting variations in the field of view, appearance discrepancy, and image quality~\cite{ting2019artificial}.
	Wang \etal~\cite{wang2019patch} have} shown that the segmentation performance of CNNs degrades rapidly on a new dataset from an unseen domain with a distribution discrepancy to the training datasets.
Since it is impossible to pre-collect all the types of datasets {for} training,
developing generalizable CNN-based fundus image segmentation algorithms on unseen datasets is a particularly pressing and challenging topic.
In this paper, we aim to effectively {explore the} knowledge {extracted} from multi-source domains to enhance the generalization of CNNs on the dataset from the unseen target domain.
As shown in Fig.~\ref{fig:background}, given annotated fundus images from multiple source domains, we expect to train a segmentation network{,} which can generalize well to new images from {an} unseen target domain, {\textit{without the need of accessing the target domain images in the training}}.

To alleviate the performance degradation of {the} trained networks in different domains, researchers have explored various methods for medical image analysis tasks.
Among them, transfer learning, \ie, fine-tuning, is a straightforward method to tackle with the domain shift of an unseen domain, but this method requires extra annotations from the new domain to fine-tune the network weights.
Considering that the annotation of medical images is expensive to acquire, some unsupervised domain adaptation methods are proposed to mitigate the distribution mismatch between {the} source and target domains~\cite{dou2018unsupervised,zhang2018task}.
Compared with transfer learning, {unsupervised domain adaptation} alleviates the demand of annotations on the target domain but still requires pre-collected target domain images in {the} training.
Moreover, it is necessary to re-train the network to adapt to a new target domain.
Unfortunately, such a setting is {highly restrictive}, as the unlabeled target domain images may not be available during the network training.
Consequently, in real clinical practice, it has broad prospects to investigate how to learn a generalizable and robust network that can be directly applied to other unseen datasets without accessing {the} target domain information or model tuning.

Recently, some related domain generalization methods are proposed {for} generalizable natural image classification~\cite{li2018domain,li2018deep,carlucci2019domain} {to narrow down the inevitable performance gap between the source and target domains}.
Nevertheless, as demonstrated in the experiment section, these methods are difficult to be extended {for} medical image segmentation problems due to the structured prediction {characteristics} of segmentation tasks.
{In medical image analysis community}, some latest methods explored various data augmentation techniques to improve the generalization ability of CNNs for medical image segmentation~\cite{chen2019improving,zhang2019unseen}.
These methods only involve one source domain dataset in training, while we aim to utilize the valuable relationship and knowledge among multiple source domains to generalize networks over {the} unseen target domains.
Very recently, Liu~\etal~\cite{liu2019large} proposed to share the visual knowledge between {the} head and tail classes via an associated visual memory to increase the recognition robustness and sensitivity for the long-tailed recognition problem.
Inspired by this memory-based meta-learning method, we propose to utilize domain knowledge memory to transfer the multi-source domain information to the unseen target domain, thereby {enhancing the generalization of segmentation networks.}

In this paper, we present a novel \textit{Domain-oriented Feature Embedding} (\textit{DoFE}) framework to improve the generalization capability of CNNs on unseen target datasets for fundus image segmentation.
Our proposed framework is able to dynamically enrich the image features with additional domain prior knowledge (\eg, domain discriminative information) learned from multi-source domains via a dynamic feature embedding mechanism.
This mechanism {facilitates} the generalization of networks in two-folds: (i) the integration between {the} semantic features of test image and domain-oriented features can be regarded as automatic feature augmentation; and (ii) it introduces extra domain prior knowledge and supplemental information to enrich the target domain image features and {to} improve their {discrimination capability}.
To this end, we incorporate a domain knowledge pool into the framework to learn and memorize the domain prior information from multi-source training datasets.
Then, we dynamically induce a domain-oriented aggregated feature from the domain knowledge pool
according to the similarity between the input test image and multi-source domain images.

Our main contributions are summarized as follows.
\begin{enumerate}
	\item[(i)] We develop a novel Domain-oriented Feature Embedding (\textit{DoFE}) framework for generalizable fundus image segmentation on unseen datasets by effectively utilizing the multi-source domain knowledge. 
	
	\item[(ii)] 
	The presented \textit{DoFE} framework is able to dynamically enrich the semantic features with an induced domain-oriented aggregated feature{, thus increasing its robustness and discrimination capability}.
	
	\item[(iii)] We design a novel {domain code} prediction branch and learning strategy to measure the similarities between {the} input test images and different {source-domain data} to facilitate {domain-oriented feature embedding}.
	
	\item[(iv)] We {evaluate} the proposed method {with} different retinal fundus image segmentation tasks, including OC/OD segmentation and vessel segmentation.
	The experiments {show} the effectiveness of our method, outperforming {previous} domain generalization and network regularization methods.
\end{enumerate}

The remainders of this paper are organized as follows.
We review the related techniques in Section~\ref{sec:relatedwork} and elaborate the proposed \textit{DoFE} framework in Section~\ref{sec:method}.
The experiments and results are presented in Section~\ref{sec:experiment}.
We further discuss our method in Section~\ref{sec:discussion} and draw the conclusion in Section~\ref{sec:conclusion}.

\section{Related Work}
\label{sec:relatedwork}

We first summarize recent works on retinal fundus segmentation.
Then, we review related techniques of enhancing the generalization ability of deep networks, including network regularization, domain adaptation, and domain generalization.
\subsection{Fundus Image Segmentation}
Fundus image structure segmentation (\ie, OC/OD segmentation and vessel segmentation) has been widely explored by extracting hand-crafted visual features and high-level CNN features.
Among hand-crafted visual features, gradient information, shape priors, texture features, and boundary information were extracted for traditional OC/OD segmentation~\cite{lowell2004optic,abramoff2007automated,joshi2011optic,cheng2013superpixel,xu2014optic}.
For the high-level CNN feature extraction, convolutional neural networks play an important role in retinal fundus image segmentation~\cite{fu2018joint, maninis2016deep, shankaranarayana2017joint,zilly2017glaucoma,sevastopolsky2017optic, fu2018disc, edupuganti2018automatic,gu2019net,zhang2019attention}.
Most of these methods attempt to improve {the} segmentation performance with more effective network architecture designs.
For example, besides a base network architecture, DRIU~\cite{maninis2016deep} {extracts} side feature maps and {employs} specialized layers to perform OD segmentation and blood vessel {segmentation}.
ResU-Net was introduced in~\cite{shankaranarayana2017joint}{; it} employs an adversarial module between the ground truth and segmentation mask to improve the final segmentation performance.
Recently, a two-stage segmentation framework {called M-Net} was presented to {jointly segment the OC and OD~\cite{fu2018joint}}.
Based on M-Net, Zhang~\etal~\cite{zhang2019attention} proposed an AG-Net with {an attention-guided module} for retinal image segmentation.
However, deep neural networks {often lack the} generalization ability and generate high test error {on unseen target datasets}, even if they have a low error rate on the training datasets~\cite{miyato2018virtual}.

\subsection{Network Regularization}

Recently, some researchers designed advanced network regularization techniques to improve the generalization capability of neural networks.
{Among the existing representative regularization approaches, Mixup trained the neural network on a linear combination of image pairs~\cite{zhang2017mixup}.
	Verma~\etal~\cite{verma2019manifold} extended Mixup to Manifold Mixup by interpolating hidden representations.}
Very recently, Yun~\etal~\cite{yun2019cutmix} presented CutMix to regularize the neural network by cutting and pasting image patches.
In another aspect, these network regularization techniques {can be} regarded as advanced data augmentation techniques on both {the} input and feature levels.

\subsection{Domain Adaptation}
In the medical image analysis field, existing domain adaptation techniques aim to find an invariant space between the source and target domains by aligning them in {the} input space~\cite{chen2018semantic,zhang2018task}, feature space~\cite{kamnitsas2017unsupervised,dou2018unsupervised}, and output space~\cite{wang2019patch}.
For {input-space} alignment, the {Cycle-GAN-based} methods were utilized to transfer the target domain images to the style-realistic source domain images~\cite{chen2018semantic,zhang2018task}.
Kamnitsas~\etal~\cite{kamnitsas2017unsupervised} performed the latent feature alignment to explore a shared feature space on the source and target domains through {the} adversarial learning.
Based on the correlated property in the output space, {output-space} adversarial learning was proposed to address the domain shift for fundus image segmentation by encouraging the segmentation in the target domain to be similar to the source ones~\cite{wang2019patch}.
Recently, the combination of {input-space} and {feature-space} alignment was also investigated in~\cite{chen2019synergistic}.
Although {domain-adaptation-based} methods mitigate domain shift problem, it requires a set of pre-collected images from the target domain to train the CNNs.

\subsection{Domain Generalization.}
Domain generalization (DG) aims to learn a universal representation from {the} source domains to improve the performance of the unseen target domain~\cite{li2018domain}.
This problem can also be regarded as a zero-shot learning problem without learning on the images or labels from the target domain~\cite{li2017deeper}.
With the development of convolutional neural networks,
Motiian~\etal~\cite{motiian2017unified} proposed an alignment loss and a separation loss to explore a {domain-invariant} feature space. 
A multi-task autoencoder was introduced in~\cite{ghifary2015domain} to transform the original image into analogs in multiple related domains to learn more robust classification features.
Similarly, a maximum mean discrepancy-based adversarial auto-encoder was presented to align the distributions among different domains~\cite{li2018domain}.
{Recently, Carlucci~\etal~\cite{carlucci2019domain} introduced an auxiliary task of solving an image jigsaw puzzle problem to enhance the generalization of the network.}
However, these works focus {on natural image classification, so} it is unclear how to extend them for medical image segmentation.
Very recently, domain generalization has drawn substantial attention in medical image segmentation.
Zhang~\etal~\cite{zhang2019unseen} proposed a deep-stacked transformations approach for medical image segmentation by combining different kinds of data augmentation.
Meanwhile, Chen~\etal~\cite{chen2019improving} also studied how to design the data normalization and augmentation techniques to improve the network generalization for medical image segmentation tasks.
However, the experiments of these methods only involve one dataset from the source domain in training, while we aim to {explore} the valuable relationship and information among datasets from multiple source domains.

\begin{figure*}[!t]
	\centering
	\includegraphics[width=0.95\linewidth]{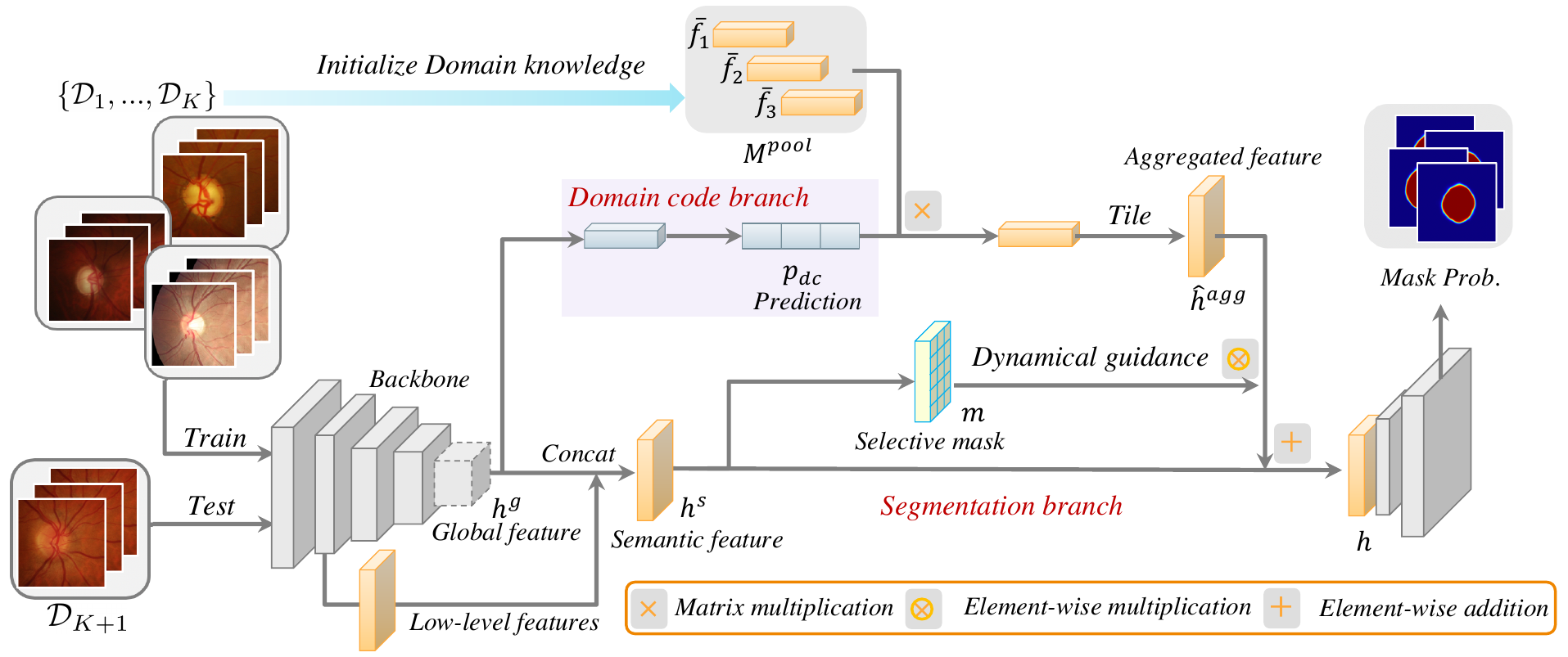}
	\caption{Schematic diagram of our proposed \textit{DoFE} framework to utilize multi-source domain datasets $\{\mathcal{D}_1, ..., \mathcal{D}_K\}$ for generalizable segmentation on the unseen dataset $\mathcal{D}_{K+1}$. 
		We {adopt} the Domain Knowledge Pool $\bm{M}^{pool}$ to learn and memorize the multi-source domain prior knowledge. 
		{Importantly, our} framework dynamically enriches the image semantic feature $\bm{h}^s$ with domain-oriented aggregated feature $\bm{\hat{h}}^{agg}$ extracted from $\bm{M}^{pool}$ to improve the {expressiveness} of the semantic feature.
	}
	\label{fig:network_arch}
	\centering
\end{figure*}

\section{Methodology}
\label{sec:method}

Medical images are often collected from different domains with {distribution shifts}.
Formally, we consider a set of $K$ datasets $\{\mathcal{D}_1, ..., \mathcal{D}_K\}$ with corresponding annotations from $K$ different source domains and an extra dataset $\mathcal{D}_{K+1}$ from {an} unseen target domain.
As shown in Fig.~\ref{fig:network_arch}, we present a novel domain-oriented feature embedding (\textit{DoFE}) framework to improve {its} generalization ability {on dataset} $\mathcal{D}_{K+1}$ from the target domain, without utilizing any information of $\mathcal{D}_{K+1}$ during the network training.
Our proposed framework dynamically enriches {semantic feature} $\bm{h}^s$ of {the} input image with {domain-oriented} aggregated feature $\bm{\hat{h}}^{agg}$.
Therefore, the domain prior knowledge learned from the multi-source domains could contribute to {the} target image segmentation during the inference phase.
In the following subsections, we will elaborate the domain-oriented feature embedding and the learning strategy of the whole network.

\subsection{Domain-oriented Feature Embedding}
To segment the fundus images from {the} target domain, one straightforward method is {to train a vanilla} segmentation network in a unified fashion by {directly feeding multi-source domain images into the network}.
{Although such a vanilla network may generalize} well over multi-source domains, the {learned semantic features still lack} sufficient {expressiveness} for the images from the unseen target domain, as the network did not receive supervised information from {the} images {from} the unseen target domain.
We propose to {integrate \textit{a memory module}~\cite{liu2019large,santoro2016meta} into the vanilla network to improve its generalizable capability.}
To be specific, we introduce a domain knowledge pool $\bm M^{pool}$ into the segmentation framework to learn and memorize the multi-source domain prior knowledge, as shown in Fig.~\ref{fig:network_arch}.
Then we induce domain-oriented aggregated feature $\bm h^{agg}$ from this knowledge pool for each input image by considering the similarity between the input image and multi-source domain images.
The original semantic {feature} $\bm{h}^s$ is dynamically augmented with induced aggregated {feature} $\bm{\hat{h}}^{agg}$ to acquire the final embedded feature $\bm{h}$ for further {fine-grained} semantic segmentation.

\subsubsection{Domain knowledge pool}
To preserve and utilize the {multi-source-domain} knowledge during the feature embedding, we explicitly incorporate domain knowledge pool $\bm M^{pool}$ into the network.
Within this knowledge pool, each item represents the domain prior knowledge of a single training dataset {in} source domains.
In particular, we employ the discriminative prototype of each source domain as the domain prior information. 
Formally, for the $k$-th source domain dataset $\mathcal{D}_k$, we extract high-level semantic feature $f_k^i$ (referring to Fig.~\ref{fig:network_arch}) from the pre-trained segmentation network for each input image $x_k^i \in \mathcal{D}_k$.
We then calculate the average semantic feature along the spatial dimension
\begin{equation}
	\label{eq:pool}
	\bar{f}_k = \frac{1}{N_k\times H \times W} \sum_{j}^{H\times W}\sum_{i}^{N_k}f_k^{i,j}
\end{equation} 
as the initialization of {the} $k$-th item in the domain knowledge pool, where $N_k$ denotes the number of input images in the $k$-th dataset $\mathcal{D}_k$, and $H$, $W$ represent the height and width of the feature map $\bm h^s$, respectively.
Besides calculating the average feature, other techniques can also be used to combine different image features in one domain to acquire the domain prior knowledge. However, in our experiments, we empirically found that using this average operation is almost effortlessly learned, and it considers both the intra-class compactness and inter-class dsicriminativeness. A similar strategy was also utilized in few-shot learning work~\cite{snell2017prototypical}.
{{During the training process, the domain knowledge pool is further updated alongside the training with momentum, to find a more discriminative representation of each domain, following:}
	\begin{equation}
		\label{eq:pool_update}
		\bar{f}_k = \lambda \bar{f}_k + (1-\lambda) \frac{1}{B_k\times H \times W} \sum_{j}^{H\times W}\sum_{i}^{B_k}f_k^{i,j}
	\end{equation} 
	where $B_k$ is the image number of domain $k$ in one mini-batch, and $\lambda$ is the momentum to control the update speed and set as $0.9$.
}

We employ DeepLabV3+~\cite{chen2018encoder} with {the} MobileNetV2~\cite{sandler2018mobilenetv2} backbone as the basic segmentation framework, as shown in Fig.~\ref{fig:network_arch}.
Due to the importance of low-level features for semantic segmentation task, we concatenate the low-level feature with the high-level global feature $\bm h^g$ for further {fine-grained} segmentation.
And we calculate the domain prior knowledge from the concatenated feature $\bm h^s$.
Specifically, the low and high-level features are extracted before the \textit{ReLU} and Batch Normalization layers~\cite{ioffe2015batch} to keep the value distribution.
After concatenation, the features are further normalized to constrain the distribution {to follow} a Gaussian distribution with zero mean and unit variance.
Then we utilize the normalized features to initialize the domain knowledge pool.

\subsubsection{Domain similarity learning}
For a given input image, we aim to compose domain-oriented aggregated feature $\bm h^{agg}$ from the domain knowledge pool $\bm M^{pool}$ to enrich its semantic feature $\bm h^s$ {to be more discriminative}.
As the input image may be more similar to {images from} a certain source domain than others, there exist different relationships among all the domains. 
This means that the domain prior information has different contribution values during the $\bm h^{agg}$ calculation.
We propose to automatically learn this relationship by a novel {domain code} prediction branch along with the segmentation network.

Specifically, since the {domain code} relates more to the high-level feature  information, we add the {domain code} prediction branch following the high-level global feature $\bm{h}^g$.
The {domain code} prediction branch consists of a global average pooling layer, a Batch Normalization layer, a \textit{ReLU} activation layer, and a convolutional layer. 
The last convolution layer is employed for final {domain code} prediction ($p_{dc}$), and we further use a \textit{Softmax} activation layer to normalize the predicted values.
{\color{black}This {domain code} prediction branch is optimized by the ``domain attribute" of the training dataset (\ie, which domain the training image belongs to) from the multi-source domains.}
It can be applied to distinguish {which domain the training images come from during} the training phase, and estimate the domain similarity during the testing phase.

\subsubsection{Domain-oriented aggregated feature}

The domain-oriented aggregated  feature $\bm h^{agg}$ is formulated as {a} weighted sum of the items in {the} domain knowledge pool according to the {domain code} (referring to Fig.~\ref{fig:network_arch}):
\begin{equation}
	\bm h^{agg} = \bm p_{dc} \times \bm M^{pool},
\end{equation}
where $\times $ represents the matrix multiplication, and $\bm p_{dc} \in \mathbb{R}^{K}$, $\bm M^{pool} \in \mathbb{R}^{K \times {H} \times {W} \times {C}}$, and $\bm h^{agg} \in \mathbb{R}^{{H} \times {W} \times {C}}$ (We ignore the dimension of {the} batch size here). 
And ${C}$ represents the values of channel number.
Then we tile feature $\bm h^{agg}$ into $\bm{\hat{h}}^{agg}$ with the shape of $\bm h^s$ for further calculation.

\subsubsection{Dynamic feature embedding}
In this step, we dynamically augment the original semantic feature $\bm{h}^s$ with the tiled domain-oriented aggregated feature $\bm{\hat{h}}^{agg}$ with {an attention-guided} mechanism.
In this way, the aggregated features could be selected dynamically and increase the variety and discrimination of {the} feature embedding in our problem setting.
Specifically, we add a convolutional layer following the original semantic feature $\bm{h}^s$ and then use a \textit{tanh} activation layer to generate {self-attention} map $\bm{m}$ (selective mask):
\begin{equation}
	\bm{m} = tanh(conv(\bm{h}^s)).
\end{equation}
The final dynamic domain-oriented feature $\bm{h}$ is represented as
\begin{equation}
	\bm{h} = \bm{h}^s + \bm{m}\otimes \bm{\hat{h}}^{agg},
\end{equation}
where $\otimes $ represents element-wise multiplication and $+$ {represents} the element-wise addition.
We feed the domain-oriented feature $\bm{h}$ into the following components of the segmentation network to generate {the fine-grained} segmentation masks.

\begin{table*} [!th]
	\centering
	\caption{Statistics of the public fundus image datasets used in our experiments.}
	\label{tab:datasetstatistics}
	\begin{tabular}
		{c|c|l|c|l}
		\toprule[1pt]
		{\color{black}{Task}} &    Domain No. & Dataset &  \# samples (train+test) & Scanners  \B \\
		\hline
		\multirow{4}{*}{{\color{black}{OC/OD segmentation}}} & Domain 1 & Drishti-GS \cite{sivaswamy2015comprehensive}  &  50 + 51 & (Aravind eye hospital)  \T \B \\
		\cline{2-5} & Domain 2 & RIM-ONE-r3 \cite{fumero2011rim} &99 + 60 & Nidek  AFC-210 \T \B\\
		\cline{2-5} &Domain 3 & REFUGE \cite{orlando2020refuge} (train) &  320 + 80 & Zeiss Visucam 500 \T \B \\
		\cline{2-5} &Domain 4 & REFUGE \cite{orlando2020refuge} (val) &  320 + 80  & Canon CR-2 \T \\
		\toprule[1pt]
		\multirow{4}{*}{{\color{black}{Vessel segmentation}}} & {\color{black}{Domain 1}} & {\color{black}{DRIVE \cite{staal2004ridge}}}  &  {\color{black}{20 + 20}} &{\color{black}{Canon CR5 nonmydriatic 3CCD}}   \T \B \\
		\cline{2-5} & {\color{black}{Domain 2}} & {\color{black}{HRF \cite{budai2013robust}}} &{\color{black}{15 + 30}} & {\color{black}{Canon CR-1}} \T \B\\
		\cline{2-5} &{\color{black}{Domain 3}} & {\color{black}{STARE  \cite{hoover2000locating}}} &  {\color{black}{10 + 10}} &  {\color{black}{TopCon TRV-50}} \T \B \\
		\cline{2-5} &{\color{black}{Domain 4}} & {\color{black}{CHASE-DB1 \cite{fraz2012ensemble}}} &  {\color{black}{20 + 8}}  & {\color{black}{Nidek NM-200D}} \T \\
		\toprule[1pt]
	\end{tabular}
\end{table*}

As mentioned above, this domain-oriented feature embedding scheme plays different roles in the training and testing {phases}.
During the training {phase}, this scheme can be seen as a kind of automatic feature augmentation, which regularizes the segmentation network to learn more representative features.
While in the testing phase, we explicitly incorporate the knowledge from multi-source training domains to recalibrate the feature embedding of {the} target domain images to increase {the} discrimination and representation abilities {of the network}.
Thus, our feature embedding approach could produce more general and discriminative features for both source and target domain images.

\subsection{Learning Strategy}
\label{sec:learning}
In this subsection, we introduce how to optimize the {domain code} prediction branch and the whole framework in detail.
\subsubsection{Domain code smooth}
The {domain code} can {indicate} which domain the training input image belongs to.
Therefore, we formulate the domain code prediction learning as a classification problem and use the {``domain attribute'' of the} source domain images to optimize the domain-code prediction branch. 
Since all the images from the same domain have the same domain code, the direct domain code regression would be prone to {over-fit the} training domain images.
To alleviate this problem, we design a domain code smooth strategy by randomly smoothing the hard one-hot domain code. 
More specifically, for the $i$-th input image from the $k$-th source domain, the hard one-hot ground truth ($y_{dc}^{k,i}$) for the predicted domain code can be represented as
\begin{equation}
	y_{dc}^{k,i}= [s_1, s_2, ..., s_k, ..., s_K],
\end{equation}
where $s_k=1$ and {all the} other items equal to zero.
We smooth the hard one-hot ground truth by randomly {perturbing} $s_k$ into the range $[0.8, 1.0]$ and assigning random {non-negative} values to {the} other items {while} satisfying the following constraint{:}
\begin{equation}
	\sum_{i=1}^{K}s_i=1, \quad where \ s_k \in [0.8,1.0].
\end{equation}
Then, the training of the {domain code} prediction branch can be regarded as a regression problem. 
We use the Mean Square Error (MSE) as the training objective function, which can be represented {as}
\begin{align}
	\mathcal{L}_{dc}\  =& \ \frac{1}{N} \sum_{k}^{K} \sum_{i}^{N_k} (p_{dc}^{k,i}-y_{dc}^{k,i})^2,\\
	where& \ \ N = N_1+N_2+...+N_K \nonumber,
\end{align}
where $\mathcal{L}_{dc}$ {denotes the domain code} classification loss,
and $p_{dc}^{k,i}$ and $y_{dc}^{k,i}$ are the predicted domain code and smooth ground truth of the $i$-th image from $k$-th source domain, respectively. And $N$ is the total number of images from all source domains.

\subsubsection{Total objective function}
For the segmentation loss, we employ the binary Cross-Entropy loss (BCE) for {OD} and {OC} segmentation following the multi-label setting in \cite{wang2019patch}. Then the supervised segmentation loss can be optimized by 
\begin{equation}
	\mathcal{L}_{s} \ = \ \frac{1}{N}\sum_{k}^{K} \sum_{i}^{N_k} [y_s^{k,i} \text{log} (p_s^{k,i}) + (1-y_s^{k,i})\text{log}(1-p_s^{k,i})].
\end{equation}
To optimize the whole framework, we combine the domain code prediction loss $\mathcal{L}_{dc}$ and supervised segmentation loss $\mathcal{L}_{s}$.
Then the total objective function to train the whole framework is defined as
\begin{equation}
	\mathcal{L}\  = \ \mathcal{L}_{s} + \alpha \mathcal{L}_{dc},\ \ 
\end{equation}
where $\alpha$ is the balanced weight, which is empirically set to $0.1$ in our experiments.

\subsubsection{Training strategy}
{We first train a vanilla segmentation network without the proposed domain-oriented feature embedding module.}
{We then utilize this trained segmentation network to extract the domain-specific semantic feature $\bm h^s$ to initialize the domain knowledge pool according to Eq.~\eqref{eq:pool} and train the whole framework.}
During the training process, we randomly choose the training images from the multiple source domains to form a training batch to increase {the} diversity.
The fed training images are regarded {as pseudo} test images to learn the domain similarity and update the domain knowledge pool.
Then the embedded features of images from the target domain could {enhance} segmentation results.

\section{Experiments}
\label{sec:experiment}

\subsection{Datasets and Experiment Setting}
\begin{figure}[!t]
	\centering
	\includegraphics[width=0.75\linewidth]{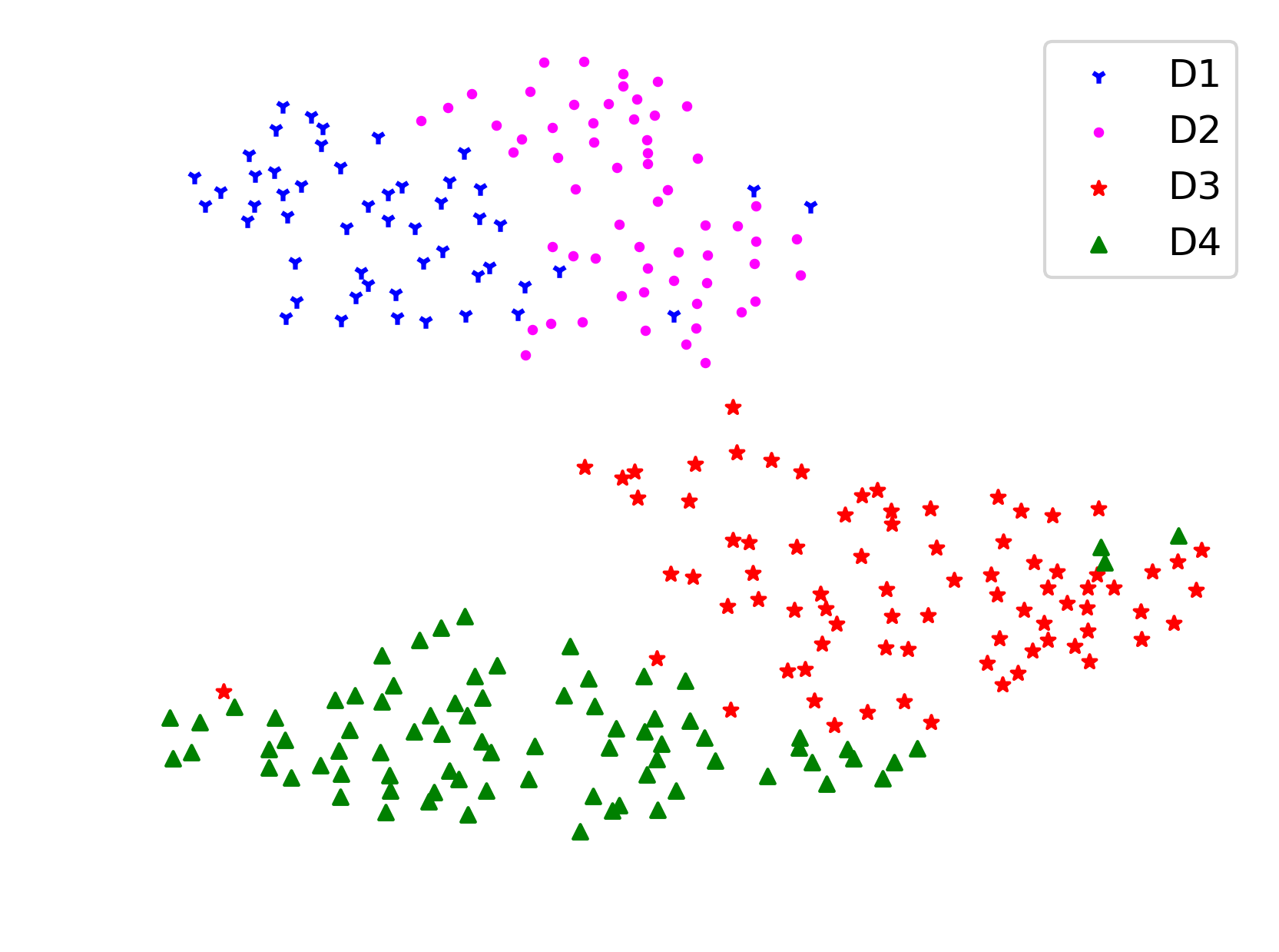}
	\caption{{{t-SNE visualization of features (extracted by a VGG16 network trained on ImageNet) of images from four datasets for OC/OD segmentation. We use different colors to denote different datasets: D1, D2, D3, and D4.}}
	}
	\label{fig:tsne}
	\centering
\end{figure}

We evaluate our approach on two segmentation tasks of retinal fundus images: OC/OD segmentation and vessel segmentation.
For each task, we conduct experiments on four public fundus image datasets{, which are captured with different scanners in different institutions.}
The detailed statistics of each dataset are shown in Table~\ref{tab:datasetstatistics}.
{Here,} we refer each dataset as a certain domain { and show the domain discrepancies of different datasets} in appearance and image quality {in Figs.}~\ref{fig:background} and~\ref{fig:result}.
{ For the OC/OD segmentation datasets, we utilized t-SNE \cite{maaten2008visualizing} to visualize the image features from different domains, where the features are extracted from the ImageNet pre-trained VGG16 network \cite{simonyan2014very}.}
{As shown in Fig.~\ref{fig:tsne}, we use different colors to denote} the image features from different datasets.
From the visualization, we can see that the {image features of different datasets} are more or less separated from one another.
{For the OC/OD segmentation task, the images from Domains 1 and 2 are split into training and testing sets following the dataset providers, while for the images from Domain 3 and 4, we randomly partition them into training and testing sets based on a ratio of $4:1$.}
{\color{black}{For the vessel segmentation task, we split the images following previous literature \cite{fu2016deepvessel,yan2018joint}.}}
{In our experiment setting, we take turns to choose each dataset as the target domain to evaluate the method performance} and {use} the remaining three datasets as {the} multi-source domains {to train the network}. 
Note that only the training image sets in {the} multi-source domains are fed into the network for training, and only the testing image sets in the target domain are utilized to evaluate {the} network performance.

\subsection{Evaluation Metrics}
{\color{black}{We adopt three metrics to evaluate the OC/OD segmentation performance, including Dice Similarity Coefficient ($DSC$), 95\% of the Hausdorff Distance ($HD$), and Average Surface Distance ($ASD$).
		The definition of $DSC$ is represented as
		\begin{gather}
		DSC  = \frac{2 \times {TP}}{{2 \times {TP} + {FP} + {FN}}},
		\end{gather}
		where $TP$, $TN$, $FP$, and $FN$ denote the numbers of true positives, true negatives, false positives, and false negatives, respectively, at the pixel level.
		{We calculate $HD$ and $ASD$ following} {the} MedPy library\footnote{\url{https://pypi.org/project/MedPy}}.
		For the vessel segmentation task, we employ accuracy ($ACC$), specificity ($SP$), sensitivity ($SE$), and the area under the receiving operator characteristic ($ROC$) curve ($AUC$) as the evaluation metrics. The definition of $ACC$, $SP$, and $SE$ are represented as
		\begin{gather}
		ACC  = \frac{TP+TN}{N}, \ \ \ SP  = \frac{TN}{TN+FP}, \\ \nonumber
		and \ \ SE  = \frac{TP}{TP+FN},
		\end{gather}
		where $N$ is the total number of testing samples.
}}
Due to the randomness of the network training, we {ran} each experiment three times and report the average and standard deviation of three predictions.

\begin{table*} [!t]
	\centering
	\caption{{Quantitative comparison with recent network generalization and  domain generalization methods on the OC/OD segmentation task [\%]. The results of Domain 1 are from the model trained with images from the other three domains. We ran each experiment three times and report the mean performance and standard deviation. 
			Top results are highlighted in \textbf{bold}.}}
	\label{tab:results}
	\resizebox{1.0\textwidth}{!}{
		\setlength\tabcolsep{1.5pt}
		\begin{tabular}{l|c|c|c|c|c|c|c|c|c}
			\toprule[1pt]
			\multirow{2}{*}{\textbf{Method}} & \multicolumn{2}{c|}{\textbf{Domain 1}} & \multicolumn{2}{c|}{\textbf{Domain 2}} & \multicolumn{2}{c|}{\textbf{Domain 3}} & \multicolumn{2}{c|}{\textbf{Domain 4}} & \textbf{Average} \B \\
			\cline{2-10} & $DSC_{cup}$  & $DSC_{disc}$   &  $DSC_{cup}$  & $DSC_{disc}$ & $DSC_{cup}$  & $DSC_{disc}$   &  $DSC_{cup}$  & $DSC_{disc}$  & $DSC$  \T \\
			\hline  
			\T \B
			Baseline &  $77.03\pm0.66$ &$94.96 \pm0.33 $  & $ 78.21\pm1.23 $& $89.69\pm1.58 $ 
			&  $80.28\pm4.03$ &$89.33 \pm1.49 $  & $ 84.74 \pm1.13 $& $90.09\pm3.67 $       
			& $85.54 $  \\  \hline \T \B
			18'mixup~\cite{zhang2017mixup} & $73.32\pm1.21$ &$92.97 \pm0.36 $  & $ 71.22\pm5.37 $& $86.78\pm2.47 $& $82.16\pm2.49$ &$90.42 \pm0.22 $  & ${{86.23}}\pm0.59 $& $90.76\pm0.57 $
			& $84.23$ \\  \T \B
			19'M-mixup~\cite{verma2019manifold} & $79.27\pm1.96$ &$94.48 \pm0.56 $  & $ 75.41\pm1.65 $& $89.38\pm0.83 $ & $83.01\pm0.72$ &${\color{black}{92.17}} \pm0.31 $  & $ 86.73\pm0.16 $& $90.82\pm1.11 $ & $86.41$ \\  \T \B
			19'CutMix~\cite{yun2019cutmix} & $76.97\pm3.64$ &$93.83 \pm 1.05 $  & $ {\color{black}{\textbf{81.02}}}\pm3.48 $& ${\color{black}{\textbf{91.97}}}\pm0.32 $& $83.42\pm1.75$ &$90.13 \pm1.04$  & $ {\color{black}{{86.83}}}\pm0.55 $& $88.79\pm1.06 $& $86.62 $ \\ \hline
			\T \B
			19'DST~\cite{zhang2019unseen} & $75.63\pm2.53$ &$92.20\pm0.61$  & $ {\color{black}{{80.80}}}\pm0.26$& ${\color{black}{90.77}}\pm0.52$& ${\color{black}{84.32}}\pm0.83$ &${\color{black}{\textbf{94.02}}}\pm0.17$  & $86.24\pm1.31$& $90.66\pm0.72$ & ${\color{black}{86.83}}$ \\ 
			\T \B
			19'JiGen~\cite{carlucci2019domain} & ${\color{black}{{80.81}}}\pm3.95$ &${\color{black}{{95.03}}}\pm0.23$  & $ 79.46\pm2.57$& ${90.47}\pm1.38$& $82.65\pm1.71$ &$91.94\pm2.17$  & $84.30\pm2.18$& ${\color{black}{{91.06}}}\pm0.30$ & $86.97$ \\  
			\hline 
			\T \B
			\textbf{\textit{DoFE}} (Ours) & ${\color{black}{\textbf{83.59}}}\pm0.19$ &${\color{black}{\textbf{95.59}}} \pm0.22$  & $ {\color{black}{80.00}}\pm0.71 $& $89.37\pm0.54$& ${\color{black}{\textbf{86.66}}}\pm0.51$ &$91.98 \pm0.24 $  & $ {\color{black}\textbf{87.04}}\pm0.48 $& ${\color{black}{\textbf{93.32}}}\pm0.33  $ & ${\color{black}{\textbf{88.44}}}$ \\
			\hline 
			\toprule[1pt]
		\end{tabular}
	}
\end{table*}

\subsection{Implementation Details}
The framework was implemented in Python based on {the} PyTorch \cite{paszke2017automatic} platform.
We used the Adam optimizer to train the whole framework, and the weights of the backbone network {was} initialized with the weights trained on the ImageNet dataset \cite{deng2009imagenet}.
{We first pre-trained the vanilla DeepLabV3+ network for 40 epochs with a learning rate of $1e-3$ and then trained the whole framework for another 80 epochs with an initial learning rate of $1e-3$. The learning rate was then decreased to $2e-4$ after 60 epochs.}
We ran the experiments on one NVIDIA TITAN Xp GPU with a batch size of 16.
{For the OC/OD segmentation task,} we cropped $800\times 800$ ROIs centering at OD by utilizing a simple U-Net~\cite{ronneberger2015u} and then resized them to $256\times 256$ as the network input.
We used the basic data augmentation to expand the training samples, { including} random scaling and cropping.
{
	We conducted the morphological operation, \ie, filling the hole, to post-process the predicted masks.}

\begin{table*} [!h]
	\centering
	\caption{{Quantitative comparison with recent domain generalization methods on the Hausdorff Distance (HD) and Average Surface Distance (ASD) [pixel] metrics for OC/OD segmentation. Top results are highlighted in \textbf{bold}.}}
	\label{tab:results_HD}
	{
		\setlength\tabcolsep{1.5pt}
		\begin{tabular}{l|c|c|c|c|c|c|c}
			\toprule[1pt]
			
			\textbf{{\color{black}{Method}}} & \textbf{{\color{black}{Metric}}} & \textbf{{\color{black}{Structure}}} & \textbf{{\color{black}{Domain 1}}} & \textbf{{\color{black}{Domain 2}}} & \textbf{{\color{black}{Domain 3}}} & \textbf{{\color{black}{Domain 4}}} & \textbf{{\color{black}{Average}}} \T \\ \hline
			
			\multirow{4}{*}{{\color{black}{19'JiGen~\cite{carlucci2019domain}}}} & \multirow{2}{*}{{{\color{black}{$HD \downarrow$}}}} & {{\color{black}{OC}}} & {\color{black}{$35.42\pm5.08$}} & {\color{black}{$25.63\pm2.49$}} & {\color{black}{$23.74\pm1.62$}} & {\color{black}{$18.59\pm2.80$}} & {\color{black}{$25.85$}} \T \B \\ 
			& & {{\color{black}{OD}}} & {\color{black}{$16.54\pm0.35 $}} & {\color{black}{$ 24.14\pm3.19$}} & {\color{black}{$23.20\pm3.31$}} & {\color{black}{$28.15\pm6.70$}} & {\color{black}{$23.01$}}\T \B \\
			\cline{2-8}&\multirow{2}{*}{{{\color{black}{$ASD \downarrow$}}}}&{{\color{black}{OC}}} & {\color{black}{$19.56\pm3.93$}} & {\color{black}{$13.99\pm1.73$ }} & {\color{black}{$11.90\pm0.98$}} & {\color{black}{$8.92\pm1.51$}} & {\color{black}{$13.60$}} \T \B \\
			& & {{\color{black}{OD}}} & {\color{black}{$8.55\pm0.40$}} & {\color{black}{$14.09\pm2.22$}} & {\color{black}{$11.35\pm2.71$}} &{\color{black}{$12.57\pm1.75$}} & {\color{black}{$11.64$}} \T \B \\ \hline \toprule[0.7pt]
			
			\multirow{4}{*}{{\color{black}{19'DST~\cite{zhang2019unseen}}}} & \multirow{2}{*}{{{\color{black}{$HD \downarrow$}}}} & {{\color{black}{OC}}} & {\color{black}{$ 43.89\pm4.24$}} & {\color{black}{${24.85}\pm0.72$}} & {\color{black}{$21.73\pm0.50$}} & {\color{black}{${14.69}\pm0.78$}} & {\color{black}{$26.29$}} \T \B \\ 
			& & {{\color{black}{OD}}} & {\color{black}{$21.84\pm0.91$}} & {\color{black}{${22.83}\pm1.07$}} & {\color{black}{${17.43}\pm0.12$}} & {\color{black}{$17.95\pm0.77$}} & {\color{black}{$\textbf{20.01}$}} \T \B \\
			\cline{2-8}&\multirow{2}{*}{{{\color{black}{$ASD \downarrow$}} }}&{{\color{black}{OC}}} & {\color{black}{$24.42\pm1.96$}} & {\color{black}{${12.89}\pm0.99$}} & {\color{black}{$10.91\pm0.48$}} & {\color{black}{${7.05}\pm0.53$}} & {\color{black}{$13.82$}} \T \B \\
			& & {{\color{black}{OD}}} & {\color{black}{$13.24\pm1.02$}} & {\color{black}{${14.00}\pm0.21$}} & {\color{black}{${8.52}\pm0.23$}} & {\color{black}{$10.05\pm0.75$}} & {\color{black}{$11.45$}} \T \B \\ \hline \toprule[0.7pt]

			\multirow{4}{*}{{\color{black}{\textbf{\textit{DoFE}} (Ours)}}} & \multirow{2}{*}{{{\color{black}{$HD \downarrow$}}}} & {{\color{black}{OC}}} & {\color{black}{${33.56}\pm0.37$}} & {\color{black}{$25.86\pm0.92$}} & {\color{black}{${19.44}\pm0.44$}} & {\color{black}{${15.76}\pm0.28$}} & {\color{black}{$\textbf{23.66}$}} \T \B \\ 
			& & {{\color{black}{OD}}} & {\color{black}{${16.21}\pm0.29 $}}&
			{\color{black}{$ 30.23\pm2.00$}} & {\color{black}{${21.45}\pm0.28$}} & {\color{black}{${16.10}\pm0.99$}} & {\color{black}{${21.00}$}} \T \B \\
			\cline{2-8}&\multirow{2}{*}{{{\color{black}{$ASD \downarrow$}}}} & {\color{black}{{OC}}} & {\color{black}{$ {16.94}\pm0.28 $}} & {\color{black}{$ 13.87\pm0.97$}} & {\color{black}{${9.59}\pm0.33$}} & {\color{black}{${7.24}\pm0.18$}} & {\color{black}{$\textbf{11.91}$}} \T \B \\
			& & {{\color{black}{OD}}} & {\color{black}{${7.68}\pm0.35 $}} & {\color{black}{$16.59\pm0.40$}} & {\color{black}{${11.19}\pm0.32$}} & {\color{black}{${7.53}\pm0.39$}} & {\color{black}{$\textbf{10.75}$}} \T \B\\
			
			\hline 
			\toprule[1pt]
		\end{tabular}
	}
\end{table*}

\subsection{Experimental Results on OC/OD Segmentation}

\subsubsection{Comparison with the baseline model}
For the baseline approach, we trained a vanilla DeepLabV3+ network with training images from all the source domains in a unified fashion.
Specifically, we treated all the source domains as one domain and trained the segmentation network in a usual supervised way. 
Table~\ref{tab:results} presents the segmentation results of the baseline model {(top row)} and our {\textit{DoFE} method (bottommost row) on the OC/OD segmentation task}.
{It is clear that our \textit{DoFE} framework surpasses the baseline model by a considerable margin (an average \textit{DSC} of 2.90\%), showing the generalization ability improvement of our \textit{DoFE} framework. 
}

\begin{figure*}[!h]
	\centering
	\includegraphics[width=0.7\linewidth]{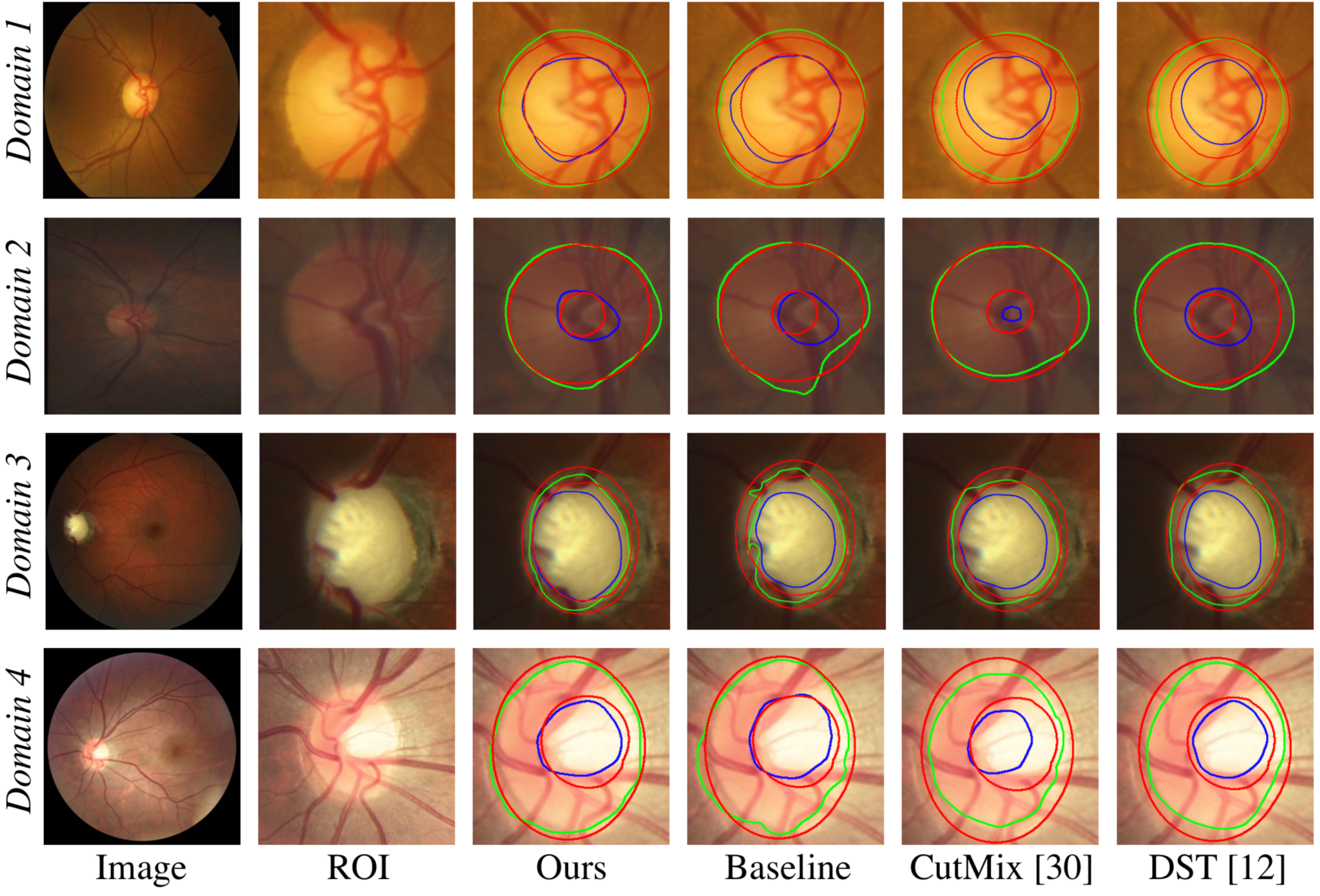}
	\caption{{Qualitative segmentation results in four domains. Each row presents one example extracted from one domain. The green and blue contours indicate the boundaries of the optic discs and optic cups, respectively, while the red contours are the ground truths. The first column shows the original images and the second column shows the enlarged region of interests (ROI).
	}}
	\label{fig:result}
	\centering
\end{figure*}

\subsubsection{Comparison with different network regularization methods}
We compare our framework with {three} recently proposed network regularization methods, including {mixup}~\cite{zhang2017mixup}, manifold mixup ({M-mixup})~\cite{verma2019manifold}, and {CutMix}~\cite{yun2019cutmix}.
We implemented the {mixup} {and {M-mixup}} by linearly {interpolating the input images and high-level features ($\bm h^s$ in our experiments), respectively.}
{We also compared with {CutMix}~\cite{yun2019cutmix} by randomly cutting patches from different domain images and then pasting them into the training images.}
As these methods were originally designed for the classification task, we employed the above operations on both images and segmentation labels to {adapt to} the segmentation task.
We referred to the public code to re-implement the above methods.
{From the results reported in Table~\ref{tab:results}}, it is observed that {mixup} does not perform well as expected in the classification task.
The average $DSC$ {of {mixup}} is worse than the baseline model.
{It may be due to the difficulty of distinguishing segmentation labels of linearly-interpolated images from fundus images that have similar appearance.}
{The {M-mixup} achieves higher $DSC$ performance than {mixup} and reaches $86.41\%$ on average.
	The reason may be that {M-mixup} applies interpolation on the feature level, which can generate more discriminative features than image-level interpolation~\cite{zhang2017mixup}.
	Further, {CutMix} achieves $86.62\%$ $DSC$ on average, $1.08\%$ higher than the Baseline.}
{Overall,} {M-mixup} and {CutMix} could preserve more structure information than {mixup} for the segmentation task,
{while these} regularization methods cannot largely { enhance} the generalization ability of the segmentation network.
Therefore, it is crucial to investigate how to boost {the} segmentation performance by utilizing the multi-source domain knowledge.

\begin{table} [!t]
	\centering
	\caption{{{Quantitative comparison on the vessel segmentation task [\%].}}
	}
	\label{tab:vessel-seg}
	\setlength\tabcolsep{1.5pt}
	\resizebox{0.5\textwidth}{!}{
		\begin{tabular}
			{c|l|c|c|c|c}
			\toprule[1pt]
			\textbf{{\color{black}Domain No.}} & \textbf{{\color{black}Method}} & {\color{black}$ACC$}  & {\color{black}$SP$}   &  {\color{black}$SE$}  & {\color{black}$AUC$} \T \\ \hline
			\multirow{2}{*}{\textbf{{\color{black}Domain 1 \cite{staal2004ridge}}}} & {\color{black}Baseline}  & {\color{black}$93.10$} &{\color{black}$95.45$}  & {\color{black}$ \textbf{68.74} $}& {\color{black}$94.09 $}  \T \B \\ 
			& \textbf{{\color{black}\textbf{\textit{DoFE}} (Ours)}} & {\color{black}$\textbf{94.11}$} &{\color{black}$\textbf{97.00}$ } & {\color{black}$64.09$}& {\color{black}$\textbf{94.23}$}          \T \B    \\ 
			\hline
			\multirow{2}{*}{\textbf{{\color{black}Domain 2 \cite{budai2013robust}}}} & {\color{black}Baseline}  & {\color{black}$83.95$} &{\color{black}$83.13$}  & {\color{black}$\textbf{93.42}$}& {\color{black}$\textbf{95.57}$}      \T \B \\ 
			& {\color{black}\textbf{\textbf{\textit{DoFE}} (Ours)}} & {\color{black}$\textbf{89.49}$} &{\color{black}$\textbf{89.48}$}  & {\color{black}$89.41$}& {\color{black}${95.47}$}         \T \B    \\ \hline
			
			\multirow{2}{*}{\textbf{{\color{black}Domain 3 \cite{hoover2000locating}}}} & {\color{black}Baseline}  & {\color{black}$91.42$} &{\color{black}$91.90$}  & {\color{black}${82.37}$}& {\color{black}$95.24$} \T \B \\ 
			&{\color{black} \textbf{\textbf{\textit{DoFE}} (Ours)} }& {\color{black}$\textbf{94.57}$} &{\color{black}$\textbf{95.27}$}  & {\color{black}$\textbf{85.53}$}& {\color{black}$\textbf{97.25}$}         \T \B    \\ \hline
			
			\multirow{2}{*}{\textbf{{\color{black}Domain 4 \cite{fraz2012ensemble}}}} & {\color{black}Baseline}  &  {\color{black}$\textbf{91.57}$} &{\color{black}$\textbf{92.34} $}  & {\color{black}$80.18 $}& {\color{black}$94.51 $}  \T \B \\ 
			& {\color{black}\textbf{\textbf{\textit{DoFE}} (Ours)}} &  {\color{black}${90.51}$} &{\color{black}${90.62}$}  & {\color{black}$\textbf{89.10}$}& {\color{black}$\textbf{96.28}$ }     \T \B    \\
			\hline
			\multirow{2}{*}{{\color{black}\textbf{Average}}} & {\color{black}Baseline}  &  {\color{black}$90.01$} &{\color{black}$90.70 $}  & {\color{black}${81.18} $}& {\color{black}$94.85 $}  \T \B \\ 
			& {\color{black}\textbf{\textbf{\textit{DoFE}} (Ours)}} &  {\color{black}$\textbf{92.17}$} &{\color{black}$\textbf{93.09}$}  & {\color{black}$\textbf{82.04}$}& {\color{black}$\textbf{95.81}$}      \T \B    \\
			\toprule[1pt]

	\end{tabular}}
\end{table}

\subsubsection{Comparison with other domain generalization methods}
So far, there is limited work {that focuses} on the domain generalization problem for the {medical image segmentation task}. Most of them {aim to design} effective data augmentation techniques. 
Among them, Chen~\etal~\cite{chen2019improving} carefully designed a data normalization and augmentation strategy to improve the generalizability of {CNNs on CMR image segmentation}.
{On the other hand, {DST} \cite{zhang2019unseen}} utilized a stack of data augmentation {strategies} for domain generalization, including random sharpening, blurring, noise, brightness adjustment, contrast change, perturbation, rotation, scaling, deformation, and cropping.
{
	Since two methods are similar, we compare our method with {DST}, which employs more data augmentation techniques. 
	From the experimental results shown in Table~\ref{tab:results}, we can observe that {DST} generates much better results and boost the average \textit{DSC} performance from $85.54\%$ to $86.83\%$, compared with the Baseline.
	Yet, our \textit{DoFE} framework still outperforms it with an average of $1.61\%$ \textit{DSC} improvement without utilizing a stack of data augmentation strategies, demonstrating the effectiveness of our domain-oriented feature embedding design.}
{Also, we compared our method with a recent domain generalization method {JiGen}~\cite{carlucci2019domain} designed for the natural image classification.}
Specifically, we added an additional branch for Jigsaw classification~\cite{carlucci2019domain} from 30 different combination types to regularize the encoder of the segmentation network.
{
	See again Table~\ref{tab:results}, it is clearly observed that our method outperforms {JiGen} by a considerable margin ($1.47\%$ average \textit{DSC}).
}
{{Besides the $DSC$ evaluation metric, we also show comparisons on $HD$ and $ASD$ with {JiGen} and {DST} in TABLE~\ref{tab:results_HD}.
		Our method generates the smallest average $ASD$ and $HD$ errors except that the $HD$ error of our method on OD segmentation is comparable with {DST}.
}}

\begin{table*} [!t]
	\centering
	\caption{Ablation study of our method [\%]. 
		{
			``w/o KP'' denotes without the knowledge pool and the feature embedding process; ``w/o DC" means that we remove the domain code prediction branch and use the average knowledge feature; ``w/o SM" means removing self-attention map guidance; and ``w/o Tr" indicates that we freeze the knowledge pool features.}
		Top results are highlighted in \textbf{bold}.}
	\label{tab:ablation_study}
	\resizebox{1.0\textwidth}{!}{
		\setlength\tabcolsep{1.5pt}
		\begin{tabular}
			{l|c|c|c|c|c|c|c|c|c}
			\toprule[1pt]
			\multirow{2}{*}{\textbf{Method}} & \multicolumn{2}{c|}{\textbf{Domain 1}} & \multicolumn{2}{c|}{\textbf{Domain 2}} & \multicolumn{2}{c|}{\textbf{Domain 3}} & \multicolumn{2}{c|}{\textbf{Domain 4}} & \textbf{Average}  \B \\
			\cline{2-10} & $DSC_{cup}$  & $DSC_{disc}$   &  $DSC_{cup}$  & $DSC_{disc}$ & $DSC_{cup}$  & $DSC_{disc}$   &  $DSC_{cup}$  & $DSC_{disc}$  & $DSC$ \T \\
			\hline  
			\T \B
			w/o KP  & $77.54\pm 5.14$ &$95.08 \pm0.21$  & $ {78.57}\pm1.35 $& $\textbf{89.62}\pm0.69 $ & $84.93\pm0.68$ &${91.26} \pm0.59 $  & $ 85.01\pm1.96 $& $91.60\pm0.86 $ & $86.86$ \\  \T \B
			w/o DC & $81.69\pm0.68$ &$95.39\pm0.04$  & $77.40\pm0.83$& $88.67\pm1.26$
			& $84.53\pm1.17$ &$\textbf{92.25}\pm0.91$  & $86.23\pm0.79$& ${91.83}\pm0.16$ 
			& $87.25$ \\  \T \B
			w/o SM & $81.48\pm1.83$ &${95.26}\pm0.06$  & $77.07\pm1.50$& $88.85\pm0.47$
			& $85.66\pm0.87$ &$91.63\pm0.84$  & $86.59\pm0.66$& $92.27\pm0.60$ 
			& $87.35$ \\   \T \B
			w/o Tr & $83.02\pm0.51$ &$\textbf{95.84}\pm0.10$  & $76.28\pm0.71$& $88.70\pm1.16$ & $85.59\pm0.47$ &$91.43\pm0.78$  & ${86.52}\pm0.34$& $92.43\pm0.91$ 
			& $87.48$ \\ \T \B
			\textbf{Ours}  & ${\textbf{83.59}}\pm0.19$ &${95.59} \pm0.22$  & $\textbf{80.00}\pm0.71 $& ${89.37}\pm0.54$& ${\textbf{86.66}}\pm0.51$ &$91.98 \pm0.24 $  & $ \textbf{87.04}\pm0.48 $& $\textbf{93.32}\pm0.33 $ & ${\textbf{88.44}}$\\
			\hline 
			\toprule[1pt]
	\end{tabular}}
\end{table*}

\subsubsection{Qualitative results}
The qualitative comparison results on {OC/OD segmentation} are further presented in Fig.~\ref{fig:result}.
As we can see, with the domain-oriented feature embedding strategy, our method generates more accurate segmentation results than the other methods.
Specifically, for the sample image in Domain 2 (the second row), it is quite hard to distinguish OD and OC for other methods due to the low image contrast, while our method is still able to segment OC and OD with accurate boundaries.
Moreover, we show an abnormal sample in the third row, where the result of our method can still outperform other methods.

\subsection{Experiment Results on Vessel Segmentation}
We present the dataset details for the vessel segmentation task in the bottom part of TABLE~\ref{tab:datasetstatistics}.
We are not aware of other domain generalization methods on the vessel segmentation task, so we directly compare our method with the vanilla DeepLabV3+ baseline to show the effectiveness of our proposed framework.
TABLE~\ref{tab:vessel-seg} reports the comparison results on the four different test domains (for each case, we take the other three datasets as the training data and train a network model). 
Our method achieves more accurate vessel segmentation on the $ACC$, $SP$, and $AUC$ metrics. 
{Note that it may not be proper to directly compare the absolute segmentation performance of our method with previous supervised methods \cite{yan2018joint,zhang2019net,gu2019net,fu2016deepvessel,zhang2019attention,maninis2016deep}, as the network backbone and the experiment settings are different.}

\begin{table} [!t]
	\centering
	\caption{{Quantitative comparison between random noise-based feature augmentation and our \textit{DoFE} framework [\%]. Our domain-oriented feature embedding surpasses the random noise-based feature augmentation.}}
	\label{tab:results_noise}
	\resizebox{0.5\textwidth}{!}
	{
		\setlength\tabcolsep{1.5pt}
		\begin{tabular}{c|c|c|c|c|c|c}
			\toprule[1pt]
			\textbf{Method} & \textbf{Structure} &\textbf{Domain 1} & \textbf{Domain 2} & \textbf{Domain 3} & \textbf{Domain 4} & \textbf{Average}  \\ \hline
			\multirow{2}{*}{\textbf{Noise}} & $DSC_{cup}$   &$82.03\pm0.86 $&$ {78.30}\pm0.70 $ &$ 84.67\pm0.51 $&$ {86.08}\pm0.79 $ &$ {82.77}$\\
			&$DSC_{disc}$   &$95.53\pm0.09 $ &${89.15}\pm0.79 $   &$91.66\pm0.59 $ &$92.87\pm0.70 $ &$92.30$\\ \hline
			\multirow{2}{*}{\textbf{Ours}} & $DSC_{cup}$   &$\textbf{83.59}\pm0.19$&$\textbf{80.00}\pm0.71$ &$\textbf{86.66}\pm0.51$&$\textbf{87.04}\pm0.48$ &$\textbf{84.32}$\\
			&$DSC_{disc}$   &$\textbf{95.59} \pm0.22$&$\textbf{89.37} \pm0.54 $  &$\textbf{91.98} \pm0.24 $&$\textbf{93.32} \pm0.33 $&$\textbf{92.57}  $\\ 
			\toprule[1pt]
		\end{tabular}
	}
\end{table}

\subsection{Analysis of Our Framework}
\subsubsection{Ablation study}
We performed thorough ablation experiments (see results in Table~\ref{tab:ablation_study}) to investigate the effect of different components in our \textit{DoFE} framework on OC/OD segmentation.
First, we removed the knowledge pool and the feature embedding components but preserved the domain code prediction branch as an auxiliary task (referred to \textit{w/o KP}).
Second, we removed the domain code prediction branch and used the average of the domain-specific prior features (referred to \textit{w/o DC}).
The third case was to remove the self-attention map $\bm{m}$ and directly combine the semantic feature $\bm{h}^s$ and domain-oriented aggregated feature $\bm{\hat{h}}^{agg}$ (referred as \textit{w/o SM}).
The last one is to investigate the influence of the training strategy, in which we fixed the features in the domain knowledge pool and did not update them during the whole training process (referred as \textit{w/o Tr}).
In this case, the knowledge pool is still initialized with the multi-source domain knowledge but without any update.
From the experimental results reported in Table~\ref{tab:ablation_study}, without the knowledge pool (\textit{w/o KP}), the average \textit{DSC} drops by $1.58\%$ to $86.86\%$ on average.
Also, the average \textit{DSC} of OC and OD on the four target datasets drops when we remove the domain code prediction (\textit{w/o DC}), the attention guided mechanism (\textit{w/o SM}), or utilize fixed knowledge information (\textit{w/o Tr}).
This ablation study illustrates the effectiveness of the major components in our framework design and the importance of dynamical domain knowledge pool update to refine the knowledge information for improved segmentation results.

\subsubsection{Analysis of domain-oriented feature enhancement}
Our method shares a similar spirit with feature augmentation to some extent.
To investigate the advantage of domain knowledge, we replaced the proposed domain knowledge pool with random noise for feature augmentation. 
Specifically, we removed the dynamic feature embedding mechanism and augmented the original semantic feature $\bm{h}_s$ with a random Gaussian noise feature {that has the same size as} $\bm{h}_s$. We conducted the same operations in both the training and testing {phases}.
The experimental results are presented in Table~\ref{tab:results_noise}.
Our full method is more accurate than using the random feature augmentation method on four datasets.
This comparison shows that the learned domain knowledge has a positive impact on the feature embedding for the segmentation task.

\begin{figure}[!t]
	\centering
	\includegraphics[width=1.0\linewidth]{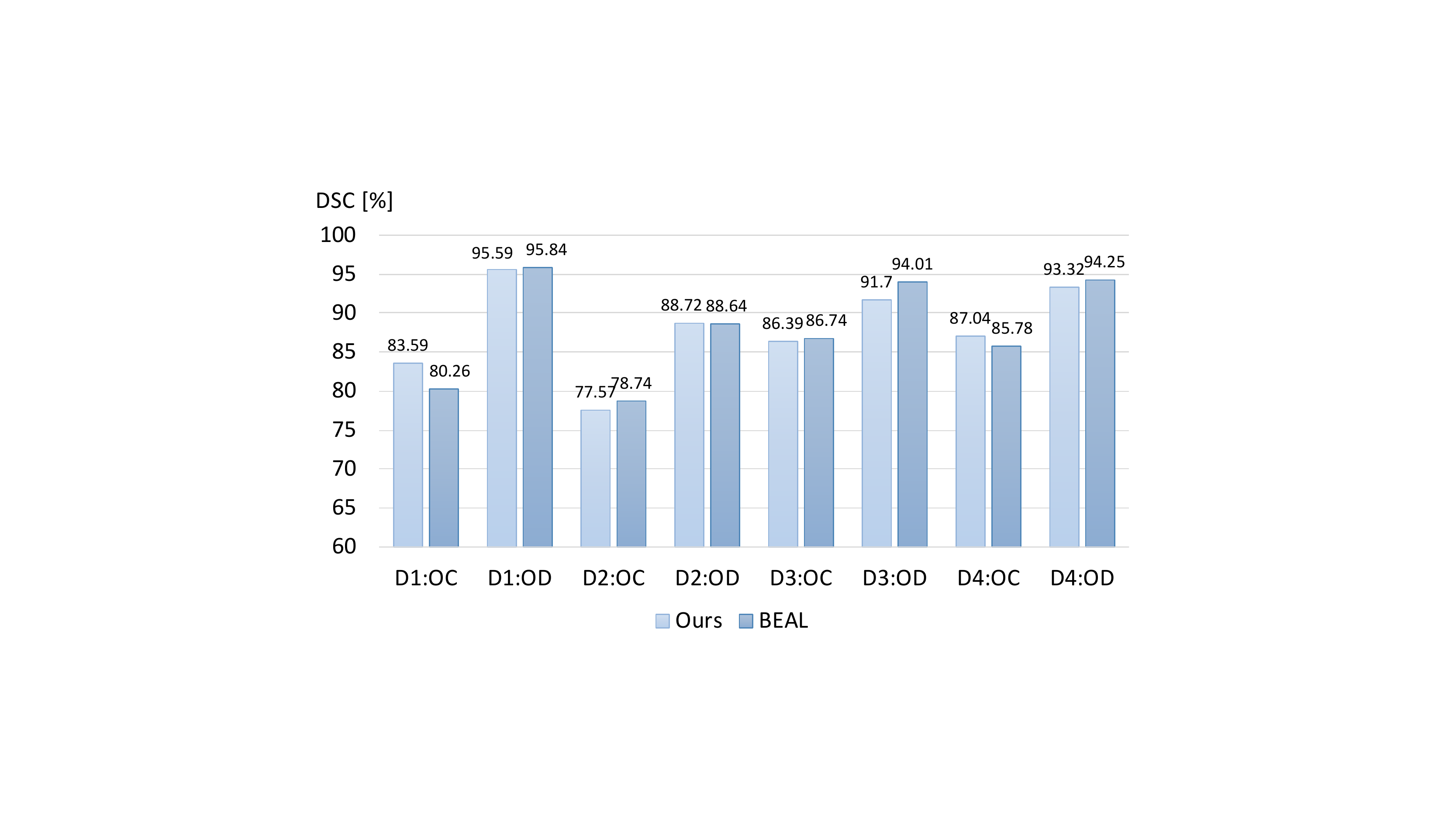}
	\caption{{Experimental results of the state-of-the-art domain adaptation method BEAL~\cite{wang2019boundary} and our framework. The x axis shows the domain number (D1/D2/D3/D4) with structure (OC/OD) and the y axis shows the dice coefficient (DSC). For example, D1:OC indicates the DSC of optic cup on target Domain 1. 
	}}
	\label{fig:results_da}
	\centering
\end{figure}

\subsubsection{{Comparison with a recent domain adaptation method}}
In our problem setting, the images from the target domain are not involved in network training.
To better evaluate the segmentation performance of our proposed method, we conducted more experiments with a domain adaptation setting, where we utilize the target domain image {during} network training.
We reproduced the state-of-the-art domain adaptation method \textit{BEAL}~\cite{wang2019boundary}.
In this experiment, the training datasets from multi-source domains are merged and regarded as the source domain, while the dataset from the target domain is regarded as the target domain for training.
The boundary and entropy information was extracted for adversarial learning to encourage the target predictions to be more similar to the source ones.
Fig.~\ref{fig:results_da} shows the comparison results.
As we can see, the domain adaptation method \textit{BEAL} achieves high performance on Domains 2 and 3, as it incorporates the extra target domain image information in network training. And the performance of our method is approaching the domain adaption method on Domains 2 and 3.
Without utilizing any target domain image in training, the performance of our method can even outperform {this recent} domain adaptation method on D1:OC and D4:OC, showing that when the training and testing domains have smaller discrepancy, our method has the potential to outperform domain adaptation methods.

\subsubsection{Statistical analysis}

To analyze the performance improvement of our method, we conducted paired \textit{t}-tests between our method and other methods, as elaborated in TABLE~\ref{tab:ttest2}.
In our paired \textit{t}-test calculation, the significance level is set as $0.05$ with a confidence level of $95\%$.
It is observed that our method has a clear improvement when compared with other methods, showing the effectiveness of our method and each designed components.

\subsubsection{Computation cost}
The computation cost of our proposed method is attractive.
The extra computation cost introduced by our method is relatively low.
The average inference time of the vanilla DeepLabV3+ is 0.70s on a single image, while our method takes about 0.73s, with only 0.03s extra time cost.
Compared with the vanilla DeepLabV3+ network architecture, our framework only adds two fully-connected layers in the domain code branch and one convolution layer in the selective mask calculation. Therefore, the additional computational cost of our method is relatively small.

\section{discussion}
\label{sec:discussion}

Convolutional neural networks have achieved promising progress in medical image analysis, {such as} retinal fundus image segmentation.
However, {one obstacle of} applying these deep-learning-based {medical image analysis} methods into real clinical practice is the limited generalization ability of {neural networks} for different datasets, which {could exhibit variations in appearance and image quality, e.g., due to the use of} various types of scanners and acquisition parameter settings~\cite{karani2018lifelong,degel2018domain,ting2019artificial}.
Therefore, researchers proposed different methods to obtain a generalizable model on unseen domains (or datasets), towards a robust medical image analysis.
To improve the generalization ability of CNN-based methods, {we aim to} avoid over-fitting on the specific training dataset and learn ``domain knowledge" from multiple source domains.
To this end, we introduced {the} Domain-oriented Feature Embedding (\textit{DoFE}) framework for fundus image segmentation {on unseen data}. 
The extensive experiments on {two public fundus image segmentation tasks} demonstrate that {our method} apparently {outperforms} other domain generalization methods and network regularization approaches.

Despite generating a superior improvement in the fundus image segmentation, {our \textit{DoFE}} still has some limitations.
{First,} multiple datasets from source domains are needed to be collected with {corresponding annotations}.
Adding more source domain datasets would certainly help to improve the overall network performance. With {an} increase in the number of source domain datasets, the network can better learn and memorize more types of images during the training. {By this means,} the network may find a better relationship between the target domain  and source domains, so that the domain-oriented aggregated features can become more discriminative.
However, annotating multiple datasets is very expensive and hard to acquire by professional ophthalmologists.
One potential solution is that we can investigate how to use unlabeled datasets from source domains to enhance the generalization ability of networks, following the spirit of semi-supervised learning~\cite{chapelle2009semi,yu2019uncertainty}.
Second, another limitation is that we consider only the general domain discrepancy between different datasets when designing our method. 
In the future work, we may further consider the inter-domain and intra-domain discrepancy problem by taking disease type and severity into count.
Furthermore,  we only conducted experiments on the fundus image segmentation tasks. We will extend our method to other medical image segmentation tasks to demonstrate the effectiveness of our proposed framework.
In the OC/OD segmentation task, the poor results often occur when the boundaries of OC and OD are gradients, which may lead to a large error. Due to the domain distribution shift,  the network cannot deal with the transition boundaries or annotation differences among images from different domains.
Currently, we only use one vector to represent the domain prior knowledge of each domain.
However, the images within one domain may have several types, leading to sub-domains in each domain.
Generating the right number of vectors in each domain can further improve the quality of the domain knowledge pool and the generalization ability of the framework.
However, it is unclear how to distinguish sub-domains in one domain dataset. Also, the number of sub-domains could be hard to determine automatically.
Exploring the feasibility of integrating unsupervised clustering \cite{caron2018deep} into our framework {could be a good direction for the future work}.

\begin{table} [!t]
	\centering
	\caption{{{Paired \textit{t}-tests between our method and others for the OC/OD segmentation task on the sample level.}}}
	\label{tab:ttest2}
	\resizebox{0.5\textwidth}{!}{
		\setlength\tabcolsep{1.5pt}
		\begin{tabular}
			{c|c|c|c|c|c}
			\toprule[1pt]
			{\color{black}\textbf{Method}} & {19'DST~\cite{zhang2019unseen}} &{\color{black}{19'JiGen~\cite{carlucci2019domain}}} & {\color{black}{19'CutMix~\cite{yun2019cutmix}}} & {\color{black}{19'M-mixup~\cite{verma2019manifold}}} &{\color{black}{18'mixup~\cite{zhang2017mixup}}}  \\
			\hline  
			\textbf{\textit{DoFE}} (Ours) & 2.05E-03 & 7.08E-08 & 1.06E-04 & 1.81E-17 & 1.08E-13 \T \\ \hline
			
			{\color{black}\textbf{Method}} & w/o KP &w/o DC& w/o SM & w/o Tr & \T \\
			\hline  
			\textbf{\textit{DoFE}} (Ours)  & 1.87E-05 & 2.13-03 & 2.04E-03 & 4.74E-03 &  \T \\ 
			\hline
			\toprule[1pt]
		\end{tabular}
	}
\end{table}

In the future, we would also like to explore how to extend our method to handle with the lifelong learning problem~\cite{parisi2019continual}, where the networks could learn and remember tasks from dynamic data distributions.
In lifelong learning, a multi-domain learner is able to incorporate new domains with only a few labeled examples (which is similar to few-shot learning), while preserving {the performance on previous domains~\cite{karani2018lifelong,baweja2018towards}}.
We will focus on improving the segmentation performance of CNN-based methods under the lifelong learning fashion.
Moreover, in our current problem setting, the target domain segmentation task must be the same as the source domains, which means that our method cannot be applied when the source and target tasks are different. 
We will study the memory-based general transfer learning methods to tackle different tasks.

\section{Conclusion}
\label{sec:conclusion}

We presented a novel Domain-oriented Feature Embedding framework for generalizable medical image segmentation by effectively utilizing the multi-source domain knowledge {learned from the domain knowledge pool}.
Specifically, we incorporated a domain knowledge pool into the framework to learn and memorize the prior domain information from multi-source domain datasets.
The knowledge features were aggregated according to domain similarity learning between the testing domain and {the} multi-source training domains{, since the} multi-source domains {could} contribute differently to the unseen domain feature embedding.
We then {designed} a novel {domain code} prediction branch along with the segmentation network to estimate these contributions.
Furthermore, we dynamically embedded the semantic features with a self-attention map.
We performed comprehensive experiments on two retinal fundus image segmentation tasks to demonstrate the improvements and effectiveness of our \textit{DoFE} framework.
More effort will be involved to extend this framework to other medical image analysis problems in the near future.

\bibliographystyle{IEEEtran}
\bibliography{ref}

\end{document}